\documentclass{article}

\usepackage[square,numbers]{natbib}

\newif\ifincludenotes
\includenotesfalse

\usepackage[preprint]{neurips_2024}

\usepackage[utf8]{inputenc} % allow utf-8 input
\usepackage[T1]{fontenc}    % use 8-bit T1 fonts
\usepackage{hyperref}       % hyperlinks
\usepackage{url}            % simple URL typesetting
\usepackage{booktabs}       % professional-quality tables
\usepackage{amsfonts}       % blackboard math symbols
\usepackage{nicefrac}       % compact symbols for 1/2, etc.
\usepackage{microtype}      % microtypography
\usepackage[table,xcdraw]{xcolor}  % For coloring cells and text
\usepackage{graphicx}
\usepackage{wrapfig}

\usepackage{booktabs}
\usepackage{multirow}
\usepackage{amssymb}
\usepackage{amsmath}
\usepackage[capitalize]{cleveref}
\usepackage{algorithm}
\usepackage{algpseudocode}

\newcommand{\draftonly}[1]{#1} 
\renewcommand{\draftonly}[1]{}
\newcommand{{\ourmethod}}{{SVAT}}
\newcommand{\githuburl}{\href{https://github.com/groundlight/vlm-visual-demonstrations}{https://github.com/groundlight/vlm-visual-demonstrations}}

\newcommand\todo[1]{\textit{\textcolor{purple}{[TODO] #1}}}

\title{Can Vision Language Models Learn from Visual Demonstrations of Ambiguous Spatial Reasoning?}

\author{%
  Bowen Zhao \\
  Groundlight \\
  Seattle, WA 98122 \\
  \texttt{bowen@groundlight.ai} \\
  \And
  Leo Parker Dirac \\
  Groundlight \\
  Seattle, WA 98122 \\
  \texttt{leo@groundlight.ai} \\
  \And
  Paulina Varshavskaya \\
  Groundlight \\
  Seattle, WA 98122 \\
  \texttt{paulina@groundlight.ai} \\
}

\begin{document}

\maketitle

\begin{abstract}
  Large vision-language models (VLMs) have become state-of-the-art for many computer vision tasks, with in-context learning (ICL) as a popular adaptation strategy for new ones. But can VLMs learn novel concepts purely from visual demonstrations, or are they limited to adapting to the output format of ICL examples? We propose a new benchmark we call Spatial Visual Ambiguity Tasks ({\ourmethod}) that challenges state-of-the-art VLMs to learn new visuospatial tasks in-context. We find that VLMs fail to do this zero-shot, and sometimes continue to fail after finetuning. However, adding simpler data to the training by curriculum learning leads to improved ICL performance. 
  % However, they can in-context learn to perform the more difficult tasks via finetuning on easier tasks through curriculum learning. 
  We release our benchmark generation, training, and evaluation code\footnote{\githuburl} to facilitate future research.
  % Honest question: What tasks are VLMs actually SOTA on?  VQA I suppose.  Increasingly, my mental model is that VLMs have a really bad inductive bias for a lot of pure CV tasks.
\end{abstract}

\section{Introduction}

% Gist: ICL is popular for adapting VLMs to new domains but does it work for novel visual concepts?
Pretrained large vision language models (VLMs) have become essential tools and set new state-of-the-art in many general-purpose vision tasks~\citep{chen2024far,Lin_2024_CVPR,Liu_2024_CVPR,yao2024minicpm}. Extensive pretraining data allow VLMs to operate in novel domains without fine-tuning, either zero-shot, or with few-shot in-context learning (ICL)~\citep{NEURIPS2023_398ae57e,zhao2024mmicl,zhou2024visual}. However, as spatial information can be ambiguous in language~\citep{Yang_2021_CVPR}, it remains unclear what it takes to get VLMs to learn a novel visuospatial concept from visual demonstrations.   

% Gist: sometimes text explanations are ambiguous in visual referent, and could be supplemented with disambiguating images, but only if VLMs can use that visual info.
We focus specifically on the ambiguity of visual referent in the text input to the VLMs, as AI-naive users of computer vision systems in novel domains may assume background knowledge or context that the VLMs would be missing~\citep{laurenccon2024building}. For example, the word ``fiducial'' in a novel industrial domain could refer to any number of markings on a piece of equipment to be aligned, and can only be disambiguated with context. Including visual information in the form of labeled ICL examples with images should lead to the desired disambiguation, but only if VLMs are able to correctly analyze the information within the example images. Existing research has demonstrated that large language models only learn the task's expected output format described in the ICL examples~\citep{min-etal-2022-rethinking}. Recent work has also probed VLMs and found them incapable of solving straightforward tasks that specifically require visual information processing, where answers cannot be guessed from text alone~\citep{rahmanzadehgervi2024vision}. 

In this paper, we explore how this combination of VLM and ICL limitations prevents quick adaptation of VLMs to novel tasks where the core concept of the task is introduced in the vision modality, and the query text is ambiguous. Specifically, we propose a new benchmark for ambiguous visual-spatial tasks called Spatial Visual Ambiguity Tasks ({\ourmethod}). It is a set of tasks of varying degrees of difficulty, where each task is to identify the correct spatial decision boundary in a synthesized image based on very limited ambiguous text and a number of visual demonstrations. Degrees of difficulty are achieved by varying the complexity level of the objects in the foreground and the image background, as well as the number of distracting objects (ambiguous visual referents) present in the image. 

We evaluate state-of-the-art VLMs through tasks in {\ourmethod} in three settings: zero-shot, directly finetuned, or finetuned through a curriculum learning (CL)~\citep{10.1145/1553374.1553380} approach.
Our experiments show that state-of-the-art VLMs fail at tasks in {\ourmethod} in the zero-shot setup without finetuning. While simply finetuning VLMs on {\ourmethod} can boost their performance by 5.8\%-27.3\% across different models, we show that curriculum learning {\ourmethod} enables VLMs to achieve better accuracy on the most challenging {\ourmethod} task, with 14.2\% to 34.2\% relative accuracy gains compared to direct fine-tuning.

\section{{\ourmethod} Benchmark}
\begin{figure}[htbp]
    \centering
    \begin{minipage}[b]{0.3\textwidth}
        \centering
        \includegraphics[width=\textwidth]{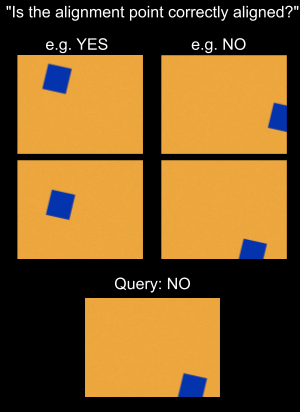}
        \textbf{(a)} Trivial background
    \end{minipage}
    \hfill
    \begin{minipage}[b]{0.3\textwidth}
        \centering
        \includegraphics[width=\textwidth]{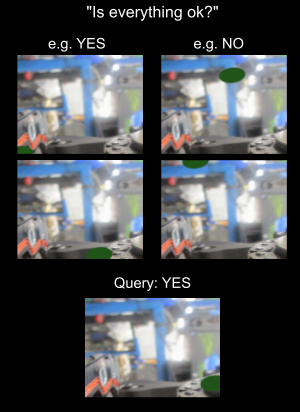}
        \textbf{(b)} Complex background
    \end{minipage}
    \hfill
    \begin{minipage}[b]{0.3\textwidth}
        \centering
        \includegraphics[width=\textwidth]{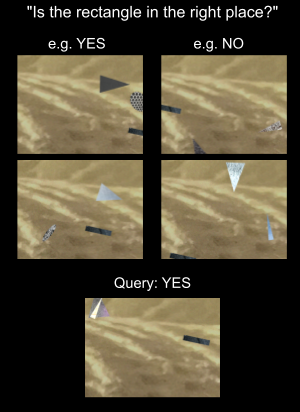}
        \textbf{(c)} Distracting objects
    \end{minipage}
    \caption{Examples of {\ourmethod} tasks where the object of interest is a simple shape.  In (a) the colors and textures are trivial with $\varphi = (\mathbb{I}_2, \mathbb{C}_{\text{shape}}, 1, \mathbb{T}_{\text{none}})$, while in (b) there is more visual complexity with $\varphi = (\mathbb{I}_5, \mathbb{C}_{\text{shape}}, 1, \mathbb{T}_{\text{none}})$.  In (c) there are distractor shapes, and the model must identify the object of interest using the text of the query, with $\varphi = (\mathbb{I}_3, \mathbb{C}_{\text{tshape}}, 3, \mathbb{T}_{\text{guide}})$}
    \label{fig:sample-tasks}
\end{figure}

We propose the {\ourmethod} benchmark to study the capabilities of VLMs on ambiguous visual-spatial reasoning through ICL. The benchmark consists of a series of classification tasks, where the goal is to learn from in-context examples the correct region of the image in which an object of interest should be located. Task difficulty is varied by the information provided in the text input, the complexity of the object of interest, the number of distracting objects present in the image, and the complexity of the image background. As intuitive examples show in \cref{fig:sample-tasks}, the core task in {\ourmethod} is to answer whether a foreground object is present within the image's ``correct'' location. The unusual challenge is that the``correct'' location is not explicitly defined but must be inferred by the model using the in-context examples. In detail, \cref{sec:background} introduces the general background of visual ICL tasks, \cref{sec:problem-formulation} formally describes how {\ourmethod} fits to visual ICL task families, \cref{sec:dataset} presents the dataset construction algorithm, and \cref{sec:curriculum} details the curriculum learning (CL) setup we use to improve VLMs' performance on {\ourmethod}.

% For example, in \ref{fig:sample-tasks}(a) the two YES examples have the rectangle in the left side of the image, while the two NO examples have the rectangle in the right half of the image, so we can reasonably infer that the decision boundary is a horizontal line dividing the top and bottom halves of the image.  The text query in this case (``Is the alignment point correctly aligned?'') provides no useful information whatsoever, but keeps the task consistent with other VQA problems.

% In more difficult versions of the task, the text query is necessary to figure out which is the object of interest.  For example, in \ref{fig:sample-tasks}(c) there are multiple shapes in different positions, but the query text says ``Is the rectangle in the right place?''  For this task, the model must use the text "rectangle" to know which of the shapes to look for in order to identify the decision boundary.  The additional objects on the image are referred to as distracting objects, or distractors.  In all cases, the final answer is simply YES or NO. \todo{reduce the contents in these leading paragraphs}

% Images come from here: https://docs.google.com/presentation/d/1lIUU5AUbOSOnL_lJA7C61dodiCQ2YbD9JGjLc8oWTWc/edit#slide=id.g3002012768c_0_122

\ifincludenotes
    \todo{This table will be removed for publication}
    \begin{itemize}
        \item \textbf{Example $e$:} an individual text, image, label tuple, $e = (t, v, y)$.  
        \item \textbf{Image $v$:} the image in the tuple $e$, consisting of a background image $i$ and a number of foreground objects $o_i = (c_i, \xi_i)$
        \item \textbf{Input $q$:} a set of N Examples, which all share the same text.  But one of the labels is held out / unobserved.
        \item \textbf{task family:} A generative process for queries $\mathbb{E}_{\varphi}$, parameterized by $\varphi$
        \item \textbf{Family parameters $\varphi$:} hyperparameters controlling difficulty of the task family
        \item \textbf{Curriculum $\mathcal{C}$}: a sequence of task families to finetune a VLM gradually towards the complex task
    \end{itemize}
\fi

\subsection{Background: Visual In-context Learning} \label{sec:background}
% \todo{Using practical examples to help demonstration; pseudo code for algorithm? Keep task formulation at a high level for Visual ICL task}
We first formulate the problem of visual ICL. Formally, under the vision question-answering (VQA) setting, we define an input prompt $x$ that consists of a set of in-context examples together with a new question and image:
\begin{equation} \label{eq:visual-icl}
    x = (E, t^q, v^q)\text{, where } E = \{e_i| e_i = (t^d_i, v^d_i, y^d_i) \in \mathbb{E}, t^d_i \in \mathbb{T}, v^d_i \in \mathbb{V}\}_{i=1}^N, t^q \in \mathbb{T}, v^q \in \mathbb{V}
\end{equation}
where $\mathbb{T}$ is a finite set of textual questions the VLM should answer, $\mathbb{V}$ is the set consisting of all possible images given a specific task, and $y_i^d$ is the ground-truth label for the image-question pair $y_i^d = l(t_i^d, v_i^d)$. Thus, a VLM is expected to tackle the task that $ y^q = \text{VLM}(E, t^q, v^q) = l(v^q, t^q)$.
% $x = (t, P, E, e^q)$ as the unit of decision-making for the model. It constitutes a natural language question $t$, a decision function $P$, a set of $N$ demonstration examples $E \subset \mathbb{E}$ and a query example $e^q = (i, {o_1, ... o_M}) \in \mathbb{E}$. Each demonstration example $e^d \in E$ and $e^q$ constitutes a background image $i$, and a list of $M$ foreground objects ${o_1, ... o_M}$. Each foreground object $o_j = (c_j, \xi_j)$, where $c$ is the class or category of the object, and $\xi$ is a low-dimensional vector that defines the pose of the object, which is all the information needed to render the foreground object onto the scene - its position, size, and orientation.

% [how the rovum task is formulated - binary QA, objects in the picture, linear decision boundary, ICL, etc.]
% Synthetic dataset. Each data is synthesized based on: object dictionary (what objects to be identified) $\mathcal{O}$, background images to be used $\mathcal{B}$, number of distracting objects $N$

% [how different difficulty levels are defined - \#distractors, background complexity, etc. This would be necessary for the coming section for curriculum learning] 

\subsection{{\ourmethod} Problem Formulation} \label{sec:problem-formulation}
{\ourmethod} fits into the visual ICL formulation in \cref{eq:visual-icl} by specifying a generation process for the text query $t$, image $v$ and label $y$. 
As shown in \cref{fig:sample-tasks}, each demonstration or query image $v = (i, o_1, ..., o_M)$ consists of a background image $i$ and $M$ foreground objects $o_1,...,o_M$, while all examples share the same text question $t$. Moreover, each foreground object $o_j = (c_j, \xi_j)$ is defined by $c$, the category of the object, and a low-dimensional vector $\xi$ that defines the pose of the object (position, size, orientation, etc.). Within each input $x$ and sharing across examples $E$ and $e^q$, we have a decision function $P$ that maps an orientation of an object $\xi$ to a label: $P : \xi \rightarrow \{0,1\}$. When there is more than one object, i.e., $M > 1$, only the first object, $o_1$ is needed to find the example's label, while the rest are left as visual distractors. Overall, an {\ourmethod} dataset $\mathcal{D}_{\text{\ourmethod}}$ is defined as:
% \begin{equation}
%     \begin{split}
%         \mathcal{D}_{\text{\ourmethod}} &= \{(x, P, y^q)|x = (E, t, v^q), y^q = l(v^q, t, P) = P(\xi_1)\} \text{, where}\\
%         E &= \{e_j| e_j = (t, v^d_j, y^d_j), y^d_j = l(v^d_j, t, P)\}_{j=1}^N, v = (i, {o_1, ... o_M}), o_k = (c_k, \xi_k)
%     \end{split}    
% \end{equation}
\begin{equation} \label{eq:svat}
    \mathcal{D}_{\text{\ourmethod}} = \{(x, P, y^q)|x = (E, t, v^q), y^q = l_{\text{\ourmethod}}(v^q, t, P) = P(\xi_1)\} 
\end{equation}
where $P$ is the decision boundary that must be inferred by the VLM from ICL examples. Note that no visual or textual clues in the image and question show $P$. The same question $t$ and decision boundary $P$ are shared across demonstration examples and the query example.
% where $l$ classifies an example given a decision boundary. In practice, in VLM training and inference, we encode the query $x$ in an image-text-interleaved manner, where the question $t$, each image example $e^d \in E$, and its label $l(e^d, P)$ is first, and then the query example $e^q$ follows at the end. \todo{Re-organize: move formal notations to the appendix, and leave the natural descriptions here}
% The VLM prompt for each data point in our benchmark dataset contains $N_E$ examples $E = \{e_1, ..., e_{N_E}\} \subset \mathbb{E}$ for ICL demonstration and a query example $q \in \mathbb{E}$ for the VLM to answer. 

\subsection{Generating {\ourmethod} Datasets} \label{sec:dataset}
In practice, the process of sampling $e = (t, v, y) \in \mathbb{E}$, especially images $v = (i, o_1, ..., o_M) \in \mathbb{V}$ is not trivial.
Since we want to examine VLMs' ambiguous spatial reasoning capabilities at different difficulty levels, {\ourmethod} should be built in a manner where the fine-grained complexity of each example is controllable, ranging from one naive shape on a solid background to a number of realistic objects on a complex photograph. Therefore, we parameterize the sampling process by $\varphi$ which comprises a set of hyperparameters related to the choice of questions, objects and images, and the nature of the decision boundary.  Each specific value of $\varphi$ defines a task family $\mathbb{E}_{\varphi} \subset \mathbb{E}$ where each example is of a similar nature and similar difficulty level.
% We have multiple competing goals at play: we want a large diversity of examples, in order to help the VLM trained on them to generalize; we want each example to contain enough information to be solvable; we want to control the difficulty of each task.  Many possible examples $e$ are intrinsically ambiguous.  For example if all the YES examples have the object of interest on the far left of the image, and all the NO examples are on the far right of the image, but the query image is in the middle, the example is clearly ambiguous.  There simply is not enough information in the in-context examples for any model to reliably decide which side of the decision boundary the query image is on.
% \todo{redo / simplify above paragraph}

% \todo{For Leo's pseudo code: given $\varphi$ and also the number of ICL examples $N$, how to construct the dataset for task family $\mathbb{E}_\varphi$}

We parameterize the difficulty $\varphi = (\mathbb{I}, \mathbb{C}, M, \mathbb{T})$ with a known set of background images $i \in \mathbb{I}$ and a known set of categories of images to be used as foreground objects $c_j \in \mathbb{C}$, as well as the number of distracting foreground objects $M$ and the set of possible text inputs $\mathbb{T}$. Text can be uninformative ($\mathbb{T}_{\text{none}}$), such as ``Is everything okay?'', or guiding the VLM ($\mathbb{T}_{\text{guide}}$) by including the name of the target object $c_1$ in the question. To avoid making {\ourmethod} tasks overly challenging, we simplify the decision boundary function $P$ to be either a horizontal or vertical line on the image $v$, while the $\xi$ for objects $o$ are two-dimensional coordinates, as shown in \cref{fig:sample-tasks}.

% The most interesting part of the generation process is that which defines the decision boundary function $P$, and the locations of each object.  In principle, $P$ could be any function, but in this study, we constrain $P$ to simple linear functions that modern VLMs have a chance to learn.  Additionally, for the experiments below $\xi$ is simply a two-dimensional location vector, keeping size and pose constant across all instances.  Furthermore, in our experiments $P$ only considers a single dimension of $\xi$, ignoring the other one.  Practically speaking this means the decision boundary is either a horizontal or a vertical line somewhere on the image.

\ifincludenotes

\begin{algorithm}
\caption{Input Prompt Generation Algorithm}
\label{alg:input-prompt-gen-english}
\begin{algorithmic}[1]
\State task $P$:= randomly select the decision boundary
\State class label $y$ := uniformly select from YES or NO
\Procedure{construct-example}{text, background image $i$, task $P$, object category $c$, class label $y_k$}
  \State randomly position one object of category $c$ into image $i$ to one side of $P$ consistent with $y_k$
  \State randomly position $M-1$ distractor objects into $i$
  \State return (text, image, $y$)
\EndProcedure
\State ICL-examples := [~]
\For {repeat $k$ times ($k=2$ in these experiments)}:
    \State add {\small CONSTRUCT-EXAMPLE}(text, $i$, $P$, $c$, YES) to ICL-examples
    \State add {\small CONSTRUCT-EXAMPLE}(text, $i$, $P$, $c$, NO) to ICL-examples
\EndFor

\State shuffle ICL-examples
\State return input prompt := ICL-examples + {\small CONSTRUCT-EXAMPLE}(text, $i$, $P$, $c$, $y$) -- $y$
\end{algorithmic}
\end{algorithm}

\fi

For all choices of $\varphi$, we keep the labels balanced.  The in-context examples always include an equal number of YES and NO examples, although the order is random.  Also, during training, the query image is equally likely to be from either class.  \cref{appendix:generation} describes the input prompt generation procedure in detail. In the experiments shown in this paper we choose $\varphi$ among five different background image sets ($\mathbb{I}_1$ to $\mathbb{I}_5$) and five foreground object category sets ($\mathbb{C}_{\text{easy}}$, $\mathbb{C}_{\text{shape}}$, $\mathbb{C}_{\text{tshape}}$, $\mathbb{C}_{\text{tool}}$, $\mathbb{C}_{\text{hard}}$), and we set the $M$ in our task families to be either 1 or 3. Thus, we curate $5 \times 5 \times 2 = 50$ task families in {\ourmethod}. Each factor ($\mathbb{I}$, $\mathbb{C}$, and $M$) would influence the difficulty level of the task to be generated. For each task family in {\ourmethod}, we generate 1,000 training, 200 validation, and 1,000 testing examples. More details of each task family's characteristics can be found in \cref{appendix:dataset-detail}.

% We have published the source code for these generating processes in \githuburl.  We hope this enables further research into learning difficult spatial reasoning tasks from in-context examples.  For the experiments reported in section \ref{sec:experiments} we use an internal proprietary set of images of tools for objects of interest ($\mathbb{C}_{\text{tool}}$ and $\mathbb{C}_{\text{hard}}$), which is not included in the open source release. However, other examples of objects, such as simple solid-colored and textured shapes ($\mathbb{C}_{\text{shape}}$ and $\mathbb{C}_{\text{tshape}}$), are included. More details of each task family's characteristics can be found in \cref{appendix:dataset-detail}

\subsection{Curriculum Learning on {\ourmethod}} \label{sec:curriculum}
The different choices of $\varphi$ form a set of task families in {\ourmethod} with varying levels of difficulty. We will show in \cref{tab:main-result} that state-of-the-art VLMs struggle to tackle complex task families, both in a zero-shot setting and after finetuning. However, progressively increasing task difficulty during CL finetuning increases VLM performance. This section formalizes CL on \ourmethod.

We define a task family $\mathbb{E}_{\varphi}$ which is a subset of all possible examples parameterized by $\varphi$, thus a curriculum $\mathcal{C}(\varphi) = (\mathbb{E}_{\varphi_1}, ..., \mathbb{E}_{\varphi_{|\mathcal{C}|})}$ is an ordered sequence of task parameterizations. Unless explicitly mentioned, we train VLMs in two stages when using CL, starting with an easier task family $\mathbb{E}_{\varphi_1}$, and then a harder task family $\mathbb{E}_{\varphi_2}$. We design four CL strategies corresponding to the three perspectives in $\varphi$ that affect the task difficulty, namely $\mathcal{C}^{\mathbb{I}}$ for background complexity, $\mathcal{C}^{\mathbb{C}}$ for object category variety, $\mathcal{C}^{M}$ for the number of distracting objects, and $\mathcal{C}^{\text{all}}$ for all aspects where we start to train VLMs from the simplest task, thus for $\varphi_i = (\mathbb{I}_i, \mathbb{C}_i, M, \mathbb{T}_i)$:
{\small
\begin{equation} \label{eq:curriculum}
    \begin{split}
        \mathcal{C}^{\mathbb{I}}(\varphi_i) &= (\mathbb{E}_{(\mathbb{I}_1, \mathbb{C}_j, M, \mathbb{T}_t)}, \mathbb{E}_{\varphi_i}), 
        \mathcal{C}^{\mathbb{C}}(\varphi_i) = (\mathbb{E}_{(\mathbb{I}_i, \mathbb{C}_{\text{easy}}, M, \mathbb{T}_t)}, \mathbb{E}_{\varphi_i}) \\
        \mathcal{C}^{M}(\varphi_i) &= (\mathbb{E}_{(\mathbb{I}_i, \mathbb{C}_j, 1, \mathbb{T}_t)}, \mathbb{E}_{\varphi_i}), 
        \mathcal{C}^{\text{all}}(\varphi_i) = (\mathbb{E}_{(\mathbb{I}_1, \mathbb{C}_{\text{easy}}, 1, \mathbb{T}_t)}, \mathbb{E}_{\varphi_i})
    \end{split}
\end{equation}
}

\section{Experiments} \label{sec:experiments}
We evaluate the capacity of VLMs to learn in-context novel visuospatial concepts in our \ourmethod\ benchmark in this section. We report the performance of several current VLMs in zero-shot, finetuned, and curriculum learning (CL) settings. We leave the discussion and limitation of {\ourmethod} in \cref{appendix:limitation}.

\subsection{Experimental Setup}

\noindent\textbf{Backbone VLMs.}
We evaluate and finetune the following VLMs pretrained on different corpora: LLaVA-Next~\citep{Liu_2024_CVPR}, VILA-1.5-8B~\citep{Lin_2024_CVPR}, Idefics2~\citep{laurençon2024matters}, InternVL2~\citep{chen2024far}, and MiniCPM-V-2.6~\citep{yao2024minicpm} from Huggingface. All of these models, except LLaVA-Next, were either pretrained on image-text-interleaved datasets (VILA and Idefics2), or are known to excel on existing multi-image benchmarks (InternVL2 and MiniCPM-V-2.6). We evaluate only the 7B (or 8B) parameter versions of each backbone for experiment efficiency and comparison fairness.
% \todo{which implementations do we use? from HF?}

\noindent\textbf{Task Selection.}
As {\ourmethod} consists of numerous task families with different selections of the parameterization $\varphi$, it would be infeasible if we enumerate every task selection throughout {\ourmethod}. Therefore, we only consider two main sets of task families $(\mathbb{I}_5, \mathbb{C}, 1, \mathbb{T}_{\text{none}})$ and $(\mathbb{I}_5, \mathbb{C}, 3, \mathbb{T}_{\text{guide}})$ in \cref{tab:main-result}, as the former one tests whether a VLM can do spatial reasoning without the help of texts, and the latter one investigates if a VLM can identify the target object based on the guided text's help. We also show the performance of VLMs on a simpler task $(\mathbb{I}_5, \mathbb{C}, 1, \mathbb{T}_{\text{guide}})$ in \cref{appendix:guide-no-distractor}, where the question explicitly mentions the target object's category but there is no distractor in the image.

\noindent\textbf{Training.}
We use ModelScope's {\it swift} library~\citep{zhao2024swiftascalablelightweightinfrastructure} to finetune VLMs on {\ourmethod}. We use LoRA~\citep{hu_lora_2021} to finetune the VLMs, either on a single task or in stages via CL. When using CL, within each difficulty level $\mathbb{E}_{\varphi_i}$, we shuffle the order of training examples and use the finetuned LoRA parameters to initialize the training for the subsequent difficulty level. After the last and most difficult finetuning step, we merge LoRA parameters with the frozen VLM backbone for evaluation. Across all experiment setups, we finetune VLMs on each task family $\mathbb{E}_{\varphi_i}$ with three epochs unless explicitly mentioned. More details of our fine-tuning setup, including hyperparameters, can be found in \cref{appendix:experiment-detail}. 
% \todo{Appendix A is one sentence. Should just add here?}

\noindent\textbf{Evaluation.} Because all {\ourmethod} tasks are simple yes/no binary tasks with 50-50 class balance, we simply report the exact-match accuracy for all tasks. Additionally we conduct one-sample z-tests on our results to see whether a VLM performs significantly better than random guessing. We set the significance level $\alpha$ as 0.05, so the threshold of any VLM performing significantly better than random guessing on each task's test set with 1,000 examples would be 52.7\%.

\subsection{Results}

% \usepackage{booktabs}
% \usepackage{multirow}
\begin{table}[t!]
\centering
\caption{Main results of VLMs' performance on {\ourmethod}. $M$ denotes the number of objects per example, and the second row on the header indicates the foreground object category set $\mathbb{C}$ in task family $\varphi$. The complexity of the background images is fixed at level 5 ($\mathbb{I}_5$). Accuracy significantly better than random guessing is in \colorbox{green!25}{green}, and each task's best model's result is in \textbf{bold}.}
\label{tab:main-result}
\resizebox{1.0\linewidth}{!}{
\begin{tabular}{@{}ll|ccccc|ccccc@{}}
\toprule
                            &               & \multicolumn{5}{c|}{$M = 1$, $\mathbb{T} = \mathbb{T}_{\text{none}}$}   & \multicolumn{5}{c}{$M = 3$, $\mathbb{T} = \mathbb{T}_{\text{guide}}$}  \\
                            &               & \multicolumn{5}{c|}{(no distractors, useless text)}   & \multicolumn{5}{c}{(distractors, text names objects)}  \\
Category                    & Model         & easy & shape & tshape & tool & hard & easy & shape & tshape & tool & hard \\ \midrule
\multirow{5}{*}{Zero-shot}  & LLaVA-Next    & 0.0  & 0.0   & 0.0    & 0.0  & 0.0  & 0.0  & 0.0   & 0.0    & 0.0  & 0.0  \\
                            & Idefics2      & 50.4 & 49.6  & 49.8   & 50.7 & 52.3 & 51.1 & \cellcolor{green!25}53.7  & \cellcolor{green!25}\textbf{52.7}   & 49.2 & 49.7 \\
                            & VILA-1.5          & 49.3 & 48.9  & 49.9   & 47.6 & 47.7 & 49.8 & 52.4  & 51.8   & \textbf{52.3} & 48.7 \\
                            & InternVL2     & 46.8 & 49.9  & 48.2   & 47.7 & 46.1 & 50.2 & \cellcolor{green!25}\textbf{54.0}  & 49.3   & 49.8 & 50.1 \\
                            & MiniCPM-V-2.6 & \cellcolor{green!25}\textbf{59.5} & \cellcolor{green!25}\textbf{57.3}  & \cellcolor{green!25}\textbf{56.5}   & \cellcolor{green!25}\textbf{58.0} & \cellcolor{green!25}\textbf{55.0} & \textbf{52.6} & 51.9  & 51.1   & 50.8 & \textbf{50.4} \\ \midrule
\multirow{5}{*}{Finetuned}  & LLaVA-Next    & \cellcolor{green!25}52.8 & 47.9  & 52.0   & 49.2 & 49.3 & \cellcolor{green!25}\textbf{80.3} & \cellcolor{green!25}53.4  & 51.1   & 49.3 & 52.3 \\
                            & Idefics2      & \cellcolor{green!25}65.6 & \cellcolor{green!25}53.9  & 51.2   & \cellcolor{green!25}54.6 & \cellcolor{green!25}62.1 & 49.0 & \cellcolor{green!25}54.1  & 50.0   & 49.7 & 48.6 \\
                            & VILA-1.5          & \cellcolor{green!25}72.9 & 49.9  & 49.9   & \cellcolor{green!25}\textbf{77.3} & \cellcolor{green!25}66.6 & 49.1 & \cellcolor{green!25}54.5  & 50.6   & 49.6 & 50.6 \\
                            & InternVL2     & \cellcolor{green!25}70.4 & \cellcolor{green!25}74.7  & \cellcolor{green!25}55.0   & \cellcolor{green!25}52.9 & 49.8 & \cellcolor{green!25}77.9 & \cellcolor{green!25}\textbf{76.9}  & 52.4   & \cellcolor{green!25}\textbf{65.6} & 50.9 \\
                            & MiniCPM-V-2.6 & \cellcolor{green!25}\textbf{73.4} & \cellcolor{green!25}\textbf{80.0}  & \cellcolor{green!25}\textbf{68.6}   & \cellcolor{green!25}74.2 & \cellcolor{green!25}\textbf{71.8} & \cellcolor{green!25}52.8 & \cellcolor{green!25}72.0  & \cellcolor{green!25}\textbf{58.4}   & 52.2 & \cellcolor{green!25}\textbf{62.1} \\ \bottomrule
\end{tabular}
}
\vspace{-1em}
\end{table}

\begin{figure}[htbp]
    \centering
    \begin{minipage}{0.64\textwidth}
        \centering
        \includegraphics[width=\linewidth]{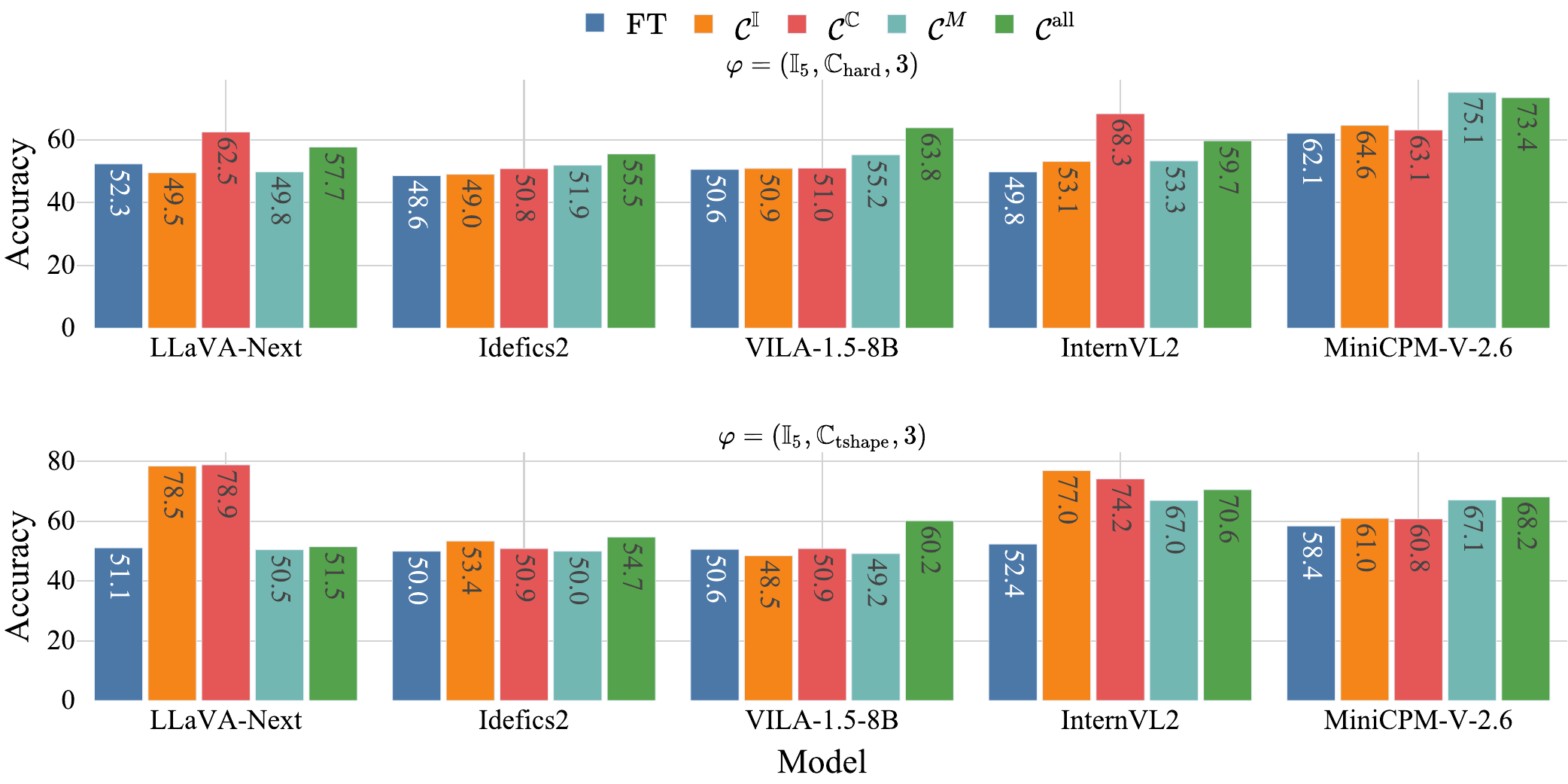}
        \caption{VLMs' performance on the task family $\varphi = (\mathbb{I}_5, \mathbb{C}_{\text{hard}}, 3, \mathbb{T}_{\text{guide}})$ and $\varphi = (\mathbb{I}_5, \mathbb{C}_{\text{tshape}}, 3, \mathbb{T}_{\text{guide}})$ after various CL strategies. FT denotes the straight-up finetuning baseline accuracy shown in \cref{tab:main-result}.}
        \label{fig:curriculum-main}
    \end{minipage}%
    \hspace{0.01\textwidth}
    \begin{minipage}{0.34\textwidth}
        \centering
        \includegraphics[width=\linewidth]{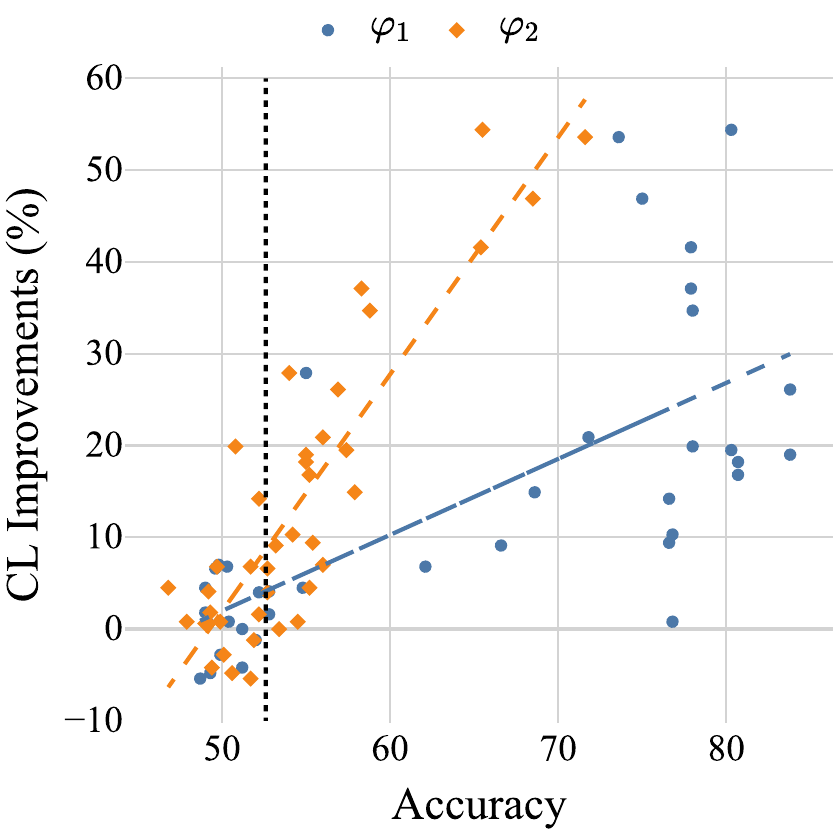}
        \caption{Correlations between the accuracy improvements after CL to the VLMs' performance on $\mathbb{E}_{\varphi_1}$ and $\mathbb{E}_{\varphi_2}$ after training with $\mathbb{E}_{\varphi_1}$.}
        \label{fig:curriculum-correlation}
    \end{minipage}
    \vspace{-2em}
\end{figure}

We demonstrate the performance of VLMs in zero-shot and finetuned settings on {\ourmethod} in \cref{tab:main-result}. In zero-shot settings, pretrained VLMs struggle with ambiguous spatial reasoning regardless of their pretraining and instruction-tuning recipes. Among the evaluated VLMs, MiniCPM performs the best across all tasks with an average accuracy of 54.3\%. It is also the only VLM that consistently achieves significantly better than random guessing on no-distractor plus useless texts tasks ($M=1$, $\mathbb{T} = \mathbb{T}_{\text{none}}$) under zero-shot settings. In the meantime, some models perform better in has-distractor, guided texts settings ($M=3$, $\mathbb{T} = \mathbb{T}_{\text{guide}}$), e.g., Idefics2 on $\mathbb{C}_{\text{tshape}}$ and InternVL2 on $\mathbb{C}_{\text{shape}}$. The reason could be that textual prompts are clearer in $M=3$, $\mathbb{T} = \mathbb{T}_{\text{guide}}$ settings where the target object's category is explicitly mentioned in the query. We conclude that these models might be more sensitive to natural language rather than vision prompts at inference. Furthermore, despite the poor accuracy of these VLMs, most models can correctly follow the format of the ICL examples to answer with either YES or NO, except for LLaVA-Next, which fails to even match random guessing performance because it does not output the correct format.  We conjecture that LLaVA-Next cannot follow multi-image ICL examples due to not being pretrained on image-text-interleaved datasets. 

\begin{wrapfigure}{r}{0.35\textwidth}
    \vspace{-1em}
    \centering
    \includegraphics[width=0.35\textwidth]{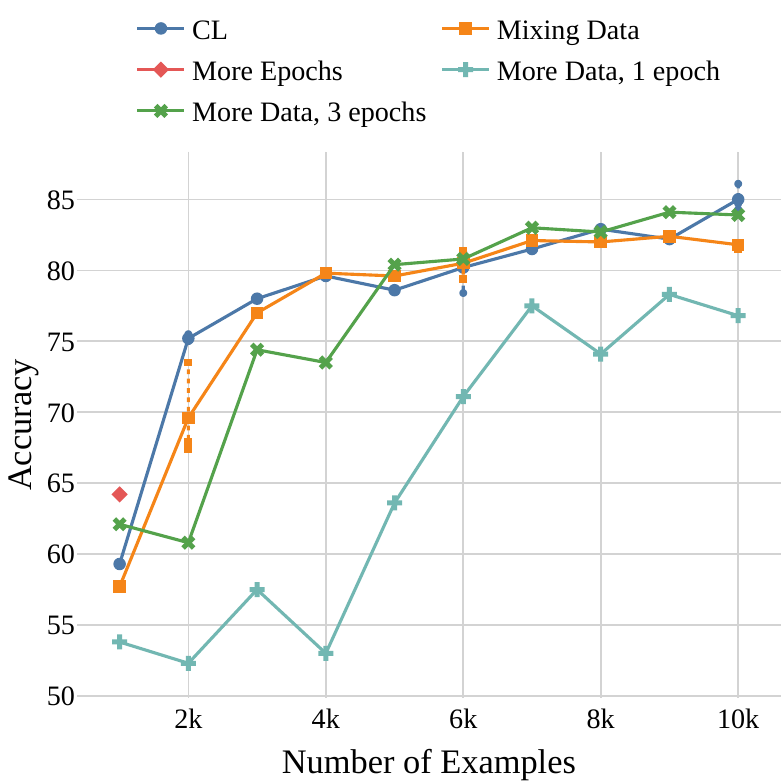}
    \caption{Ablation on MiniCPM for the task $(\mathbb{I}_5, \mathbb{C}_{\text{hard}}, 3, \mathbb{T}_{\text{guide}})$.}
    \label{fig:ablation-study}
    \vspace{-1em}
\end{wrapfigure}
% \todo{[cross-model comparisons here] [interesting to see if LLaVA-next can benefit from it] }

Finetuning VLMs directly on tasks in {\ourmethod} improves their performance (bottom section of \cref{tab:main-result}), regardless of how the model was pretrained. MiniCPM still performs best with 66.6\% accuracy on average after fine-tuning. Surprisingly, LLaVA-Next achieves extremely-high accuracy on $(\mathbb{I}_5, \mathbb{C}_{\text{easy}}, 3, \mathbb{T}_{\text{guide}})$ after finetuning, while performing poorly on $(\mathbb{I}_5, \mathbb{C}_{\text{easy}}, 1, \mathbb{T}_{\text{none}})$. We conjecture that mentioning the target object's category in the query is essential for LLaVA-Next to learn the objects' spatial relationship within the images. Similar to this phenomenon, we see that InternVL2 performs better on most $M=3, \mathbb{T}_{\text{guide}}$ tasks than their $M=1, \mathbb{T}_{\text{none}}$ counterparts. However, when the foreground object's vocabulary becomes larger (for $\mathbb{C}_{\text{tool}}$ and $\mathbb{C}_{\text{hard}}$), VLMs consistently get worse results on has-distractor settings compared to no-distractor ones. For these tasks, VLMs struggle to identify the target object even though its category is mentioned in the prompt, where only InternVL2 and MiniCPM achieve non-trivial performance on $\mathbb{C}_{\text{tool}}$ and $\mathbb{C}_{\text{hard}}$, respectively.

Although task families in {\ourmethod} with a larger object vocabulary, complex background, and some distractors are challenging for VLMs, \cref{fig:curriculum-main} shows that applying CL to VLMs effectively improves model performance. Across different models with varied curriculum setups, 34 out of 40 (85\%) trained models' performance increases compared to straightforward finetuning. Furthermore, all models can achieve significantly better accuracy after CL than random guessing ($> 52.7\%$). We also notice that different VLMs benefit most from different CL strategies. For example, $\mathcal{C}^{\mathbb{C}}$ increases the performance of LLaVA-Next and InternVL the most, while MiniCPM barely gains accuracy improvements from it. In \cref{fig:curriculum-correlation}, we show that the VLM's performance on both $\mathbb{E}_{\varphi_1}$ and $\mathbb{E}_{\varphi_2}$\footnote{We use $\mathbb{E}_{\varphi_1}$ and $\mathbb{E}_{\varphi_2}$ to represent the two tasks used in CL as defined in \cref{eq:curriculum}, i.e., $\mathcal{C}(\varphi_2) = (\varphi_1, \varphi_2)$} after the first-stage finetuning substantially affects the model's final performance on $\mathbb{E}_{\varphi_2}$ after CL. We see that all models that do not achieve significantly better accuracy on $\mathbb{E}_{\varphi_1}$ cannot improve their performance through CL. Meanwhile, a positive correlation with $R^2 = 0.77$ exists between the VLMs' $\mathbb{E}_{\varphi_2}$ performance after the first and second-stage training. We conclude that learning to tackle $\mathbb{E}_{\varphi_1}$ with spatial reasoning capabilities is necessary for succeeding on $\mathbb{E}_{\varphi_2}$ throughout CL, while the final VLM's performance of CL is predictable based on the intermediate models' performance.

\subsection{Ablation Study}
As the improvements in VLM performance can be due not (or not only) to CL but to a greater diversity of data during finetuning or a larger number of training steps, we examine these possibilities through ablations. We only apply ablations to MiniCPM for simplicity, as it is one of the most efficient VLM for training and inference among the models listed above. We run ablations on the task family $\varphi = (\mathbb{I}_5, \mathbb{C}_{\text{hard}}, 3, \mathbb{T}_{\text{guide}})$ with curriculum $\mathcal{C}^M$, where the model gains the most performance through CL. We conduct the ablation study based on the following three strategies: 
\par
\textbf{1) Mixing Data.} We naively combine and randomly shuffle the data from all the datasets in CL.
\par
\textbf{2) More Epochs.} Simply training the VLM with six epochs to match the total training steps in CL.
\par
\textbf{2) More Data.} As {\ourmethod} is a synthetic dataset, we generate more training data for the task family $\varphi = (\mathbb{I}_5, \mathbb{C}_{\text{hard}}, 3, \mathbb{T}_{\text{guide}})$. We finetune the VLM with one epoch to ensure there are no repeated examples in training to eliminate overfitting and also three epochs to match CL.

\cref{fig:ablation-study} indicates simply training VLMs with more steps on the same examples (\#examples = 1,000) does not improve the model performance. However, increasing the quantity of novel training data might help, yet the performance cannot match CL unless all examples are unique in training (\#examples = 6,000), mainly because more combinations of the target object's spatial information in demonstration and query examples are presented to the model. Meanwhile, mixing the data from easy and complex tasks can help, yet the trained VLMs' performance is slightly worse than CL. Therefore, we conclude that CL is essential for empowering VLMs on ambiguous spatial reasoning, especially in data-limited scenarios where the target complex task's training example quantity is low.

\section{Related Works}

\subsection{VLMs for Spatial Reasoning}
Despite the rapid development of vision language models (VLMs)~\citep{chen2024far,Lin_2024_CVPR,Liu_2024_CVPR,yao2024minicpm}, today's best VLMs are very limited in their ability to solve seemingly simple spatial reasoning tasks~\citep{rahmanzadehgervi2024vision,Tong_2024_CVPR}. Challenging benchmarks have been proposed to examine and improve VLMs' performance in spatial reasoning, including but not limited to spatial relationship detection~\citep{Chen_2024_CVPR,cheng2024spatialrgpt,kamath-etal-2023-whats,liu-etal-2023-visual}, object localization~\citep{Ranasinghe_2024_CVPR}, navigation~\citep{wu2024visualization}, distance measuring~\citep{cheng2024spatialrgpt}, etc. However, such tasks are delicately defined with engineered prompts so that VLMs can understand the question, yet {\ourmethod} focuses on tasks that are ambiguous in words but can be properly defined by visual demonstrations. Moreover, existing methods in tackling spatial reasoning tasks with VLMs often rely on prompt engineering~\citep{wang2024picture,wu2024visualization} and explicit spatial modeling~\citep{Banerjee_2021_ICCV,10483664}, but our analysis with {\ourmethod} finds that curriculum learning can be a more efficient way of enabling VLMs' ability in spatial reasoning.

\subsection{In-Context Learning}
% \citet{NEURIPS2022_c529dba0} showed that a transformer can be trained to fit a linear regression model to data presented as in-context demonstrations at inference time.
In-context Learning (ICL) was first introduced in~\citet{NEURIPS2020_1457c0d6}, which found that pretrained large language models can be adapted to novel tasks given several demonstration examples at inference time, rather than using them to update the model's parameters. Research found that such ICL learning process can be seen as linear regression~\citep{NEURIPS2022_c529dba0}, Bayesian models~\citep{xie2022an}, gradient descent~\citep{dai-etal-2023-gpt}, etc. However, research also pointed out that ICL might only help language models shift to a new input and output distribution rather than deeper reasoning capabilities~\citep{min-etal-2022-rethinking,wei2023larger}.

Besides language models, recent advancements in multi-image multimodal learning addressed the fact that VLMs can also learn novel tasks through ICL~\citep{Lin_2024_CVPR}. Many benchmarks have been developed to specifically examine existing VLMs' ICL capabilities with multi-image inputs~\citep{zhao2024mmicl,zong2024vl}, while recent research also stressed that ICL can be adapted to VLMs to tackle visual-related reasoning tasks~\citep{zhou2024visual}. Nonetheless, such performance improvements highly rely on the text modality of the task rather than the image modality~\citep{Baldassini_2024_CVPR}, leaving the ICL's effect in vision-oriented tasks under-explored.

\subsection{Curriculum Learning}
Curriculum learning (CL) was first proposed in \citet{10.1145/1553374.1553380}, suggesting training machine learning models from easier to harder task examples could achieve better model convergence and robustness. In the area of language modeling, CL also shows its effectiveness in both pertaining~\citep{campos2021curriculum} and finetuning~\citep{Guo_Tan_Xu_Qin_Chen_Liu_2020,xu-etal-2020-curriculum} stages. In multimodal learning, especially with VLMs, research also demonstrated that CL can improve model performance in navigation~\citep{NEURIPS2021_6f044255} and vision-language alignment~\citep{Srinivasan_2023_CVPR}.

\section{Conclusion}
We introduce a benchmark of ambiguous visual-spatial reasoning tasks, namely Spatial Visual Ambiguity Tasks ({\ourmethod}), and use it to evaluate a set of current VLMs on their ability to learn novel visuospatial concepts via in-context learning. We find that current VLMs cannot solve these tasks exclusively in context without specific training, and some still fail to learn by finetuning the tasks directly. However, they can learn the more difficult tasks from in-context visual demonstrations if they have previously been finetuned on easier tasks through a curriculum learning approach. Our analysis shows that curriculum learning presents a data-efficient and more robust way of training VLMs on {\ourmethod}. These results demonstrate more evidence of the power of curriculum learning to adapt large models. Despite the performance gained from curriculum learning, state-of-the-art VLMs require further development to reliably solve ambiguous tasks with vision prompts or demonstrations. We hope our work will facilitate future research in this direction.

\bibliographystyle{abbrvnat}
\bibliography{custom}

%%%%%%%%%%%%%%%%%%%%%%%%%%%%%%%%%%%%%%%%%%%%%%%%%%%%%%%%%%%%

\appendix

\section{Experiment Details} \label{appendix:experiment-detail}
We set the learning rate as 1e-4 for finetuning, while LoRA r and alpha are set as 8 and 32, respectively. The full hyperparameters we use for all VLMs finetuning are shown in \cref{tab:hyperparameters}.

\begin{table}[htbp]
\centering
\caption{Hyperparameters used in finetuning VLMs on {\ourmethod}}
\label{tab:hyperparameters}
\begin{tabular}{cc}
\hline
Hyperparameter                 & Value \\ \hline
Learning rate                  & 1e-4  \\
Batch size                     & 16    \\
\#Epochs                       & 3     \\
Warmup ratio                   & 0.05  \\
Weight decay                   & 0.1   \\
Optimizer                      & AdamW \\
Adam $\beta_1$                 & 0.9   \\
Adam $\beta_2$                 & 0.95  \\
Adam $\epsilon$                & 1e-8  \\
Gradient clipping              & 1.0   \\
LoRA r                         & 8     \\
LoRA $\alpha$ & 32    \\
LoRA dropout                   & 0.1   \\ \hline
\end{tabular}
\end{table}

When training and evaluating with Idefics2 models, we set \texttt{do\_image\_splitting} to True to reach the full potential of the model's capabilities. For the InternVL2 model, we leave the input image size as \texttt{(448, 448)} by default and set the maximum number of crops generated from its processor to 12. For MiniCPM-V-2.6, we set the \texttt{max\_slice\_nums} to None in both training and evaluation stages. At inference, as tasks in {\ourmethod} are all binary question answering problems that expect the VLM to respond with either ``Yes'' or ``No'', we set the \texttt{max\_new\_tokens} as 5. We do not set it to 1 because we want to see whether a pretrained VLM can directly follow the ICL demonstrations' output format in zero-shot settings (details introduced in \cref{appendix:guide-no-distractor}). Meanwhile, we do not use sampling or beam search at inference time. 

\section{Input Prompt Generation in {\ourmethod}}  \label{appendix:generation}
We demonstrate an example of full prompt in {\ourmethod} under the task family $\varphi = (\mathbb{I}_5, \mathbb{C}_{\text{hard}}, 3, \mathbb{T}_{\text{guide}})$ in \cref{tab:prompt}.
For a detailed algorithmic description of the input prompt and image generation process, see \cref{alg:input-prompt-generation}.

\begin{table}[ht]
    \centering
    \small
    \noindent\fbox{%
    \begin{minipage}{\dimexpr\linewidth\fboxsep-2\fboxrule} 
\tt 
Please answer the following question based on the provided examples.\newline
\newline
Example 1:\newline
<image>\newline
Question: Is the Heat Guns in the right position?\newline
Answer: Yes\newline
\newline
Example 2:\newline
<image>\newline
Question: Is the Heat Guns in the right position?\newline
Answer: No\newline
\newline
Example 3:\newline
<image>\newline
Question: Is the Heat Guns in the right position?\newline
Answer: No\newline
\newline
Example 4:\newline
<image>\newline
Question: Is the Heat Guns in the right position?\newline
Answer: Yes\newline
\newline
Query:\newline
<image>\newline
Question: Is the Heat Guns in the right position?\newline
Answer: 
    \end{minipage}
}

\caption{Sampled prompt from the task family $\varphi = (\mathbb{I}_5, \mathbb{C}_{\text{hard}}, 3, \mathbb{T}_{\text{guide}})$ in {\ourmethod}.}
\label{tab:prompt}
\end{table}

\renewcommand{\algorithmicrequire}{\textbf{Input:}}
\renewcommand{\algorithmicensure}{\textbf{Output:}}

\begin{algorithm}
\caption{Input Prompt Generation Algorithm}
\label{alg:input-prompt-generation}
\begin{algorithmic}[1]
\Require $\varphi = (\mathbb{I}, \mathbb{C}, M, \mathbb{T})$
\Require $N$  \Comment{Number of examples, including query.  We use $N = 5$}
\Require $\varepsilon$  \Comment{Difficulty threshold.  We use $\varepsilon = 0.05$}

% Define decision boundary P
\State $D \gets \dim(\xi)$  \Comment{Dimensionality of the pose vector. We use $D = 2$}
\State $\delta \sim \text{Uniform}\{1, \ldots, D\}$  \Comment{Pick a dimension for decision boundary}
\State $\tau \sim \text{Uniform}[2\varepsilon, 1 - 2\varepsilon]$  \Comment{Threshold for decision boundary}
\State $s \sim \text{Uniform}\{-1, 1\}$  \Comment{Direction of decision boundary}
\State $P(\xi) := I[s(\xi_\delta - \tau) > 0]$  \Comment{Define decision boundary function}
% Note: in the next version, defining the decision boundary should be a sub-function.
% But we also make use of \tau later on to tell if something is "difficult/close".
% A fix is to reformulate it to return -1, 0, 1, with -1,1 being NO/YES labels, and 0 being "close".  This would require changes in a bunch of places in the paper
% to make it consistent everywhere for -1/1.

\vspace{0.5em}
% Define shared elements across examples
\State $i \sim \mathbb{I}$  \Comment{Shared background image for all examples}
\State $t^q \sim \mathbb{T}$  \Comment{Sample text query (may be $\mathbb{T}_{none}$ or $\mathbb{T}_{guide}$)}
\State $c^* \sim \mathbb{C}$  \Comment{Sample a target object class}

\vspace{0.5em}
% Generate balanced set of labels
\State $Y_{\text{init}} \gets [0, 1] \times \lfloor N/2 \rfloor$  \Comment{Initialize balanced labels}
\State $Y_{\text{query}} \sim \text{Uniform}\{0, 1\}$  \Comment{Sample final label}
\State $Y \gets \text{Shuffle}(Y_{\text{init}}) \cup Y_{\text{query}}$  \Comment{Shuffle ICL examples}

\vspace{0.5em}
% Generate set of examples
\State $E \gets []$  \Comment{Initialize list of examples}
\For{$j = 1$ to $N$}  \Comment{Create $N$ examples}
    \State $y \gets Y[j]$  \Comment{Use pre-generated label}
    \State $O \gets \emptyset$  \Comment{Initialize set of objects for this example}
    \For{$k = 1$ to $M$}  \Comment{Create $M$ objects per example}
        \Repeat
            \State $\xi \sim \text{Uniform}[0, 1]^D$  \Comment{Sample pose}
        \Until{$P(\xi) = y$ \textbf{and} $|\xi_\delta - \tau| > \varepsilon$}
            \Comment{Check label and difficulty}
        \If{$k = 1$}
            \State $c \gets c^*$  \Comment{Use target class for first object}
        \Else
            \Repeat
                \State $c \sim \mathbb{C}$  \Comment{Sample class for distractor objects}
            \Until{$c \neq c^*$}
        \EndIf
        \State $O \gets O \cup \{(c, \xi)\}$  \Comment{Add object to example}
    \EndFor
    \State $V \gets (i, O)$ \Comment{Build the image with a background and foreground objects}
    \If{$j = N$}
        \State $v^q \gets V$  \Comment{Assign the query example's image}
    \Else
        \State $E \gets E \cup [(t^q, V, y)]$  \Comment{Add example to the demonstration list}
    \EndIf
\EndFor

\vspace{0.5em}
\State $x \gets (E, t^q, v^q)$
\State \Return $(x, P, Y_{\text{query}})$ \Comment{Finish constructing an instance in $\mathcal{D}_{\text{\ourmethod}}$}
\end{algorithmic}
\end{algorithm}

\section{Dataset Characteristics Details} \label{appendix:dataset-detail}
As described in \cref{sec:dataset}, the images in {\ourmethod} are synthesized based on different sets of background images and foreground objects which makes the difficulty of each task family $\varphi = (\mathbb{I}, \mathbb{C}, M, \mathbb{T})$ controllable. 

In detail, we have five different complexity level defined for the background images, ranging from $\mathbb{I}_1$ to $\mathbb{I}_5$, including
\begin{itemize}
    \item $\mathbb{I}_1$: empty (solid white) background;
    \item $\mathbb{I}_2$: solid background but with varied RGB colors randomly sampled from \texttt{(0, 0, 0)} to \texttt{(255, 255, 255)};
    \item $\mathbb{I}_3$: simple, realistic textured images, like grass field, snow, wood, sheet, etc.;
    \item $\mathbb{I}_4$: simple photographs taken consisting of few objects, e.g., a desk, ceiling, wall, etc.;
    \item $\mathbb{I}_5$: complex images that contain multiple realistic objects from industrial scenes. 
\end{itemize}

As for the foreground objects, we have defined the following sets:
\begin{itemize}
    \item $\mathbb{C}_{\text{easy}}$: contains five objects, including a bolt, a chain, a hardhat, a pickup truck, and a tree;
    \item $\mathbb{C}_{\text{shape}}$: consisting of five naive shapes, namely circle, pentagon, rectangle, square, and triangle. Each object is filled with a solid RGB color randomly sampled from \texttt{(0, 0, 0)} to \texttt{(255, 255, 255)};
    \item $\mathbb{C}_{\text{tshape}}$: same shapes in $\mathbb{C}_{\text{shape}}$, but filled with random textures from $\mathbb{I}_3$;
    \item $\mathbb{C}_{\text{tool}}$: a set of 87 tools commonly seen in industrial scenes, like hammer, saw, carpet knife, drill, heat gun, etc., where each category of tool has only one image;
    \item $\mathbb{C}_{\text{hard}}$: 3,437 industrial tool images from 328 categories in total.
\end{itemize}

Finally, $\mathbb{T}$ in $\varphi$ controls the construction or sampling process of questions $t$ in {\ourmethod} datasets based on the formulation in \cref{eq:visual-icl}. $t$ is built based on the following templates shown in \cref{tab:question-templates}. The \texttt{\{fiducial\}} in the template is randomly replaced with a set of synonyms (including the word ``fiducial'') if $\mathbb{T} = \mathbb{T}_{\text{none}}$, like ``marker'', ``landmark'', ``beacon'', etc. When $\mathbb{T} = \mathbb{T}_{\text{guide}}$, \texttt{\{fiducial\}} is replaced with the target object's category name $c_1$. The variable \texttt{\{description\}} in the template is randomly replaced with a set of adjectives and phrases representing the status of ``Yes'' or ``No'', like ``aligned'', ``in position'', ``out of place'', etc. Since the same question $t$ is consistent within each input $x$'s demonstration examples and query example, and all questions in {\ourmethod} task families are binary, the actual choice of the variable \texttt{\{description\}} here does not affect the ground truth label of the query example, as long as the query example's decision boundary is consistent with the demonstration examples.

\begin{table}[htbp]
\centering
\caption{Question templates in {\ourmethod}}
\label{tab:question-templates}
    \begin{tabular}{l}
        \hline
        \textbf{Templates}                                          \\ \hline
        \tt Is the \{fiducial\} \{description\}?                        \\
        \tt Are the \{fiducial\} \{description\}?                       \\
        \tt Are the \{fiducial\} \{description\}?                       \\
        \tt Can you see if the \{fiducial\} is \{description\}?         \\
        \tt Is there a problem with the \{fiducial\}?                   \\
        \tt Look at the \{fiducial\}. Is it \{description\}?            \\
        \tt Find the \{fiducial\}. Is it \{description\}?               \\
        \tt Can you see the \{fiducial\}? Is it \{description\}?        \\
        \tt Is the \{fiducial\} properly positioned?                    \\
        \tt Is the \{fiducial\} correctly aligned?                      \\
        \tt Is the \{fiducial\} in the correct position?                \\
        \tt Can you see if the \{fiducial\} is in the correct position? \\
        \tt Is the \{fiducial\} in the right place?                     \\
        \tt Find the \{fiducial\}. Is it in the right place?            \\
        \tt Can you see the \{fiducial\}? Is it in the right place?     \\
        \tt Is the \{fiducial\} in the right position?                  \\ \hline
    \end{tabular}
\end{table}

\section{Additional Results: Guided Texts without Distractors} \label{appendix:guide-no-distractor}
% \usepackage{multirow}
\begin{table}[htbp]
\centering
\caption{Zero-shot and finetuned VLMs' performance on $\varphi = (\mathbb{I}_5, \mathbb{C}, 1, \mathbb{T}_{\text{guide}})$. Accuracy significantly better than random guessing is in \colorbox{green!25}{green}, and each task's best model's result is in \textbf{bold}.}
\label{tab:guided-no-distractor}
\begin{tabular}{ll|ccccc}
\hline
 &               & \multicolumn{5}{c}{\begin{tabular}[c]{@{}c@{}}M = 1, $\mathbb{T} = \mathbb{T}_{\text{guide}}$\\ (no distractors, text names objects)\end{tabular}} \\
Category                   & Model      & easy & shape & tshape        & tool & hard \\ \hline
\multirow{5}{*}{Zero-shot} & LLaVA-Next & 0.0  & 0.0   & 0.0           & 0.0  & 0.0  \\
                           & VILA-1.5       & 14.7 & 28.4  & 26.3          & 14.2 & 20.2 \\
                           & Idefics2   & 46.5 & 49.6  & 51.8          & 50.3 & 48.4 \\
                           & InternVL2  & 48.8 & 51.0  & 50.3          & 49.2 & 49.4 \\
 & MiniCPM-V-2.6 & \cellcolor{green!25}\textbf{55.4}               & \cellcolor{green!25}\textbf{56.7}               & \textbf{52.3}               & \cellcolor{green!25}\textbf{56.0}               & \cellcolor{green!25}\textbf{53.7}              \\ \hline
\multirow{5}{*}{Finetuned} & LLaVA-Next & 46.8 & \cellcolor{green!25}76.8  & 50.9          & 50.7 & 50.2 \\
                           & VILA-1.5       & \cellcolor{green!25}70.5 & \cellcolor{green!25}53.5  & 52.5          & \cellcolor{green!25}68.3 & \cellcolor{green!25}67.5 \\
                           & Idefics2   & \cellcolor{green!25}61.3 & \cellcolor{green!25}61.8  & 51.0          & 49.9 & \cellcolor{green!25}57.3 \\
                           & InternVL2  & \cellcolor{green!25}79.3 & \cellcolor{green!25}77.8  & \cellcolor{green!25}\textbf{74.8} & \cellcolor{green!25}72.3 & \cellcolor{green!25}55.0 \\
 & MiniCPM-V-2.6 & \cellcolor{green!25}\textbf{81.7}               & \cellcolor{green!25}\textbf{81.1}               & \cellcolor{green!25}73.1                        & \cellcolor{green!25}\textbf{77.5}               & \cellcolor{green!25}\textbf{70.8}              \\ \hline
\end{tabular}
\end{table}
We demonstrate VLMs' performance on the task family $\varphi = (\mathbb{I}_5, \mathbb{C}, 1, \mathbb{T}_{\text{guide}})$ under zero-shot and finetuned settings in \cref{tab:guided-no-distractor}. Despite this task should be empirically simpler than the ones shown in \cref{tab:main-result}, we still find that VLMs struggle at tackling it under zero-shot settings, where only the MiniCPM model shows its performance significantly better than random guessing on $\mathbb{C}_{\text{easy}}$, $\mathbb{C}_{\text{shape}}$, $\mathbb{C}_{\text{tool}}$, and $\mathbb{C}_{\text{hard}}$. Besides, VILA's performance is much worse than random guessing because it does not follow the output format given in the ICL demonstrations, i.e., answering with either ``Yes'' or ``No''. 2,907 out of 5,000 answers (58.1\%) from VILA fail to follow the ICL output format. 

At the same time, after finetuning, we see that in 18 out of 25 (72\%) settings, the model performs significantly better than random guessing. MiniCPM, again, performs the best across most of the settings except for $\mathbb{C}_{\text{tshape}}$. The averaged accuracy across all models on all foreground object selection $\mathbb{C}$ after finetuning achieves 64.5, which is better than that of $(\mathbb{I}_5, \mathbb{C}, 1, \mathbb{T}_{\text{none}})$ (61.0) and $(\mathbb{I}_5, \mathbb{C}, 3, \mathbb{T}_{\text{guide}})$ (56.5) shown in \cref{tab:main-result}, indicating that prompting with VLMs with objects' category names make the task easier even if there is no distractor in the image. 

\section{Discussion and Limitations} \label{appendix:limitation}
Even though our experiment demonstrated in \cref{sec:experiments} has covered various settings in {\ourmethod}, we cannot enumerate every possible combination regarding the task parameterization $\varphi$ to examine VLMs' ambiguous spatial reasoning abilities in extreme details. However, the sample code provided at \githuburl can easily be adapted to include tasks with more combinations or choices of $\varphi$. Furthermore, the core components in {\ourmethod}, namely the decision boundary $P$, the background image $i$, foreground objects $o$, and even natural language questions $t$ can be easily extended based on our released code. We would like to leave the exploration of applying more challenging tasks on VLMs as future work.

Besides extending task variety and difficulty, this paper only examines VLMs with a scale of 7B to 8B due to the limitation of our computational capabilities. In theory, larger models have more potential for conducting visuospatial reasoning, especially under in-context learning setups. Nonetheless, we argue that models can already perform relatively well by applying curriculum learning to VLMs at the 7-8B parameter scale, with the accuracy reaching about 75\%. Therefore, a larger parameter scale might not be necessary for VLMs to do ambiguous spatial reasoning with a decent accuracy.

Finally, as foundation models get more powerful, they are increasingly good at solving real-world tasks zero-shot, without any specific training. However, for many real-world tasks, the specific nature of the goal is somewhat ambiguous, and humans struggle to clearly articulate the exact criteria necessary to define a desired outcome. Oftentimes, it is easier for a person to give examples showing "this is good" and "this is bad" than to explicitly list the exact characteristics of each example that make one good or bad. {\ourmethod} only considers ambiguous spatial reasoning tasks with synthetic data, yet no realistic data for training or evaluation is considered. We want to leave the research of combining {\ourmethod} and ambiguous, realistic multimodal data as a future direction. What knowledge can be transferred between synthetic datasets like {\ourmethod} and real-world datasets and benchmarks for VLMs remains under-explored.

\end{document}